\documentclass[11pt]{article}
\usepackage{graphicx}
\usepackage{subcaption}

\usepackage[preprint]{acl}

\usepackage[utf8]{inputenc} %
\usepackage[T1]{fontenc}    %
\usepackage{booktabs}       %
\usepackage{amsfonts}       %
\usepackage{nicefrac}       %
\usepackage{microtype}      %
\usepackage{xspace}
\usepackage{xcolor}

\usepackage{graphicx}

\usepackage{amssymb, mathtools, amsmath}
\usepackage{amsthm}

\usepackage{multirow}
\usepackage{algorithm}   
\usepackage{algpseudocode}

\usepackage{tcolorbox}

\usepackage{hyperref}
\usepackage{url}
\usepackage{xcolor}         %

\usepackage{bbm}

\usepackage{paralist}
\usepackage[inline]{enumitem}

\usepackage{rotating}

\usepackage{wrapfig}
\usepackage{multicol}

\usepackage{adjustbox}

\usepackage{subcaption}
\usepackage{pifont}
\usepackage{framed}     %
\usepackage{fancyvrb}
\usepackage{bbm}
\usepackage{booktabs}

\usepackage{listings}  %
\usepackage{array}     %
\hypersetup{
    colorlinks = true,
    citecolor=dartmouthgreen,
    linkcolor=dartmouthgreen,
    urlcolor=dartmouthgreen
}
\usepackage{listings}

\definecolor{codegreen}{rgb}{0,0.6,0}
\definecolor{codegray}{rgb}{0.5,0.5,0.5}
\definecolor{codepurple}{rgb}{0.58,0,0.82}
\definecolor{backcolour}{rgb}{0.95,0.95,0.92}

\lstdefinestyle{mystyle}{
    backgroundcolor=\color{backcolour},   
    commentstyle=\color{codegreen},
    keywordstyle=\color{magenta},
    numberstyle=\tiny\color{codegray},
    stringstyle=\color{codepurple},
    basicstyle=\ttfamily\footnotesize,
    breakatwhitespace=false,         
    breaklines=true,                 
    captionpos=b,                    
    keepspaces=true,                 
    numbers=left,                    
    numbersep=5pt,                  
    showspaces=false,                
    showstringspaces=false,
    showtabs=false,                  
    tabsize=2
}

\usepackage[framemethod=TikZ]{mdframed} %

\usepackage[capitalize,noabbrev]{cleveref}

\theoremstyle{plain}

\theoremstyle{definition}

\theoremstyle{remark}

\usepackage[greek.polutoniko, english]{babel}

\usepackage{framed}     %
\usepackage[framemethod=TikZ]{mdframed} %

\usepackage{times}
\usepackage{latexsym}

\usepackage[T1]{fontenc}

\usepackage[utf8]{inputenc}

\usepackage{microtype}

\usepackage{inconsolata}

\usepackage{graphicx}

\usetikzlibrary{arrows.meta,decorations.pathreplacing,calc,positioning,patterns,fit,backgrounds}

\usepackage{booktabs}
\usepackage{tabularx}
\usepackage{array}
\usepackage{url}

\definecolor{darkgreen}{rgb}{0.0, 0.5, 0.0}
\definecolor{blue}{rgb}{0.0, 0.47, 0.75}
\definecolor{dartmouthgreen}{rgb}{0.05, 0.5, 0.06}
\definecolor{drab}{rgb}{0.59, 0.44, 0.09}
\definecolor{navyblue}{rgb}{0.0, 0.0, 0.5}

\definecolor{darkgreen}{rgb}{0.0, 0.5, 0.0}
\definecolor{blue}{rgb}{0.0, 0.47, 0.75}
\definecolor{dartmouthgreen}{rgb}{0.05, 0.5, 0.06}
\definecolor{drab}{rgb}{0.59, 0.44, 0.09}
\definecolor{navyblue}{rgb}{0.0, 0.0, 0.5}

\newcommand{\cmark}{ {\color{dartmouthgreen} \ding{51}} }%
\newcommand{\xmark}{ {\color{red} \ding{55}} }%

\usepackage{xcolor,colortbl}

\definecolor{Gray}{gray}{0.5}
\definecolor{LightCyan}{rgb}{0.88,1,1}

\usepackage{xcolor}
\usepackage{textgreek}
\usepackage{calc} %

\newcommand{\method}{\textsc{sGPO}}

\definecolor{oiblue}{HTML}{0072B2}
\definecolor{oiorange}{HTML}{E69F00}
\definecolor{oigreen}{HTML}{009E73}
\definecolor{oipurple}{HTML}{CC79A7}
\definecolor{oisky}{HTML}{56B4E9}
\definecolor{slate}{HTML}{475569}
\definecolor{slatelt}{HTML}{94A3B8}
\definecolor{deadgray}{HTML}{9CA3AF}
\definecolor{deadfill}{HTML}{F1F5F9}
\definecolor{correctfill}{HTML}{D4F5E9}
\definecolor{wrongfill}{HTML}{FEE2E2}
\definecolor{bgorange}{HTML}{FFF7E6}
\definecolor{bggreen}{HTML}{E6F9F1}
\definecolor{bgpurple}{HTML}{F7ECF7}
\definecolor{bgsky}{HTML}{E8F4FD}
\definecolor{bgslate}{HTML}{F8FAFC}

\newcolumntype{L}{>{\raggedright\arraybackslash}X}
\newcommand{\hf}[1]{\url{#1}}

\title{\texttt{sGPO}: Trading Inference FLOPs for Training Efficiency in RLVR}

\author{Shivchander Sudalairaj \\
  AI Innovation, Red Hat \\\And
  Kai Xu \\
  AI Innovation, Red Hat \\\AND
  Akash Srivastava \\
  Core AI, IBM \\\And
  Giorgio Giannone \\
  AI Innovation, Red Hat \\
}

\begin{document}
\maketitle

\begin{abstract}
Standard Reinforcement Learning with Verifiable Rewards (RLVR) training allocates a fixed rollout budget to every query, without regard for what each query's difficulty means for the current policy. 
This leads to two symmetric failure modes: easy queries produce near-zero advantage because the policy already solves them, while unsolvable queries produce no signal because the policy never solves them. Both regimes waste training FLOPs without contributing to a learning gradient.
We introduce \emph{sorted Group Policy Optimization} (sGPO), a compute-efficient strategy that trades a small budget of inference FLOPs for a large reduction in wasted training FLOPs. The key insight is that cheap inference compute can serve as a single offline proxy for query difficulty. By generating a small batch of parallel samples per query under the initial policy, we obtain a model-aware empirical success rate. This motivates setting the training rollout group size to the inverse of this success rate, a practical rule that maximizes sample efficiency by extracting the most advantage per generated rollout.
This single profiling pass simultaneously drives data filtering (removing trivial queries and sub-sampling unsolvable ones), adaptive group size allocation, and curriculum construction (scheduling queries from easy to hard). 
sGPO matches or exceeds baseline performance while reducing total training compute by 2.5--3.1$\times$---with the upfront inference profiling cost included.
\end{abstract}

\section{Introduction}
\label{sec:intro}

Reinforcement Learning has emerged as a crucial paradigm for enhancing 
the reasoning capabilities and alignment of Large Language Models 
\citep{guo2025deepseek,jaech2024openai}. Specifically, Reinforcement 
Learning with Verifiable Rewards \citep[RLVR;][]{lambert2024tulu} and 
techniques like Group Relative Policy Optimization \citep[GRPO;][]{shao2024deepseekmath} optimize models by sampling multiple 
rollouts per query and updating the policy based on the relative 
advantage of each generated response.

\begin{figure}
    \centering
    \includegraphics[width=.9\linewidth]{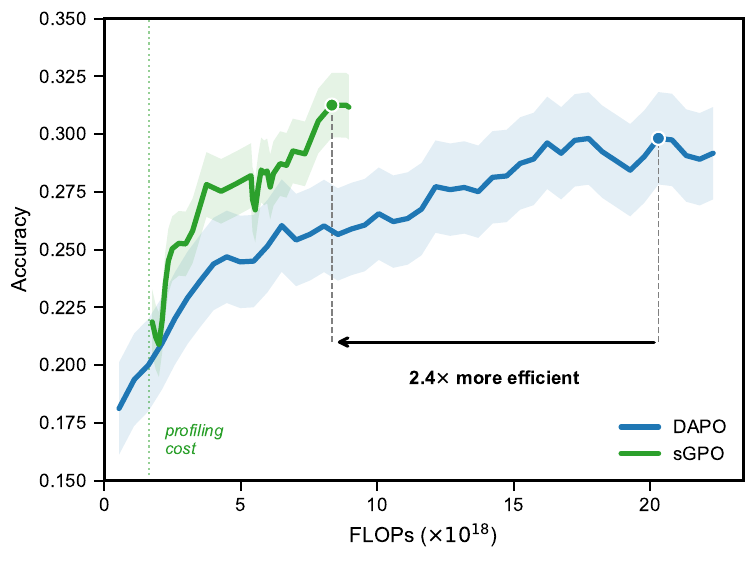}
    \caption{Accuracy-compute frontier for \method{} vs.\ DAPO on 
Qwen2.5-Math-7B. \method{} dominates throughout training, 
achieving the same accuracy at substantially lower total FLOPs. Appendix~\ref{appx:g16} for more efficiency results.
}
\label{fig:flop_efficiency}
\end{figure}

Standard RLVR training allocates a fixed rollout budget to every query 
regardless of difficulty, which leads to two symmetric failure modes. 
For trivial queries, the policy already solves them, so all rollouts 
succeed, and the relative advantage collapses to zero. For unsolvable 
queries, the policy never succeeds, again yielding no gradient. Both 
regimes waste training FLOPs without contributing a learning signal (Figure~\ref{fig:flop_efficiency}).

\begin{figure*}[ht!]
    \centering
    \begin{subfigure}[b]{0.3\linewidth}
        \centering
        \includegraphics[width=\linewidth]{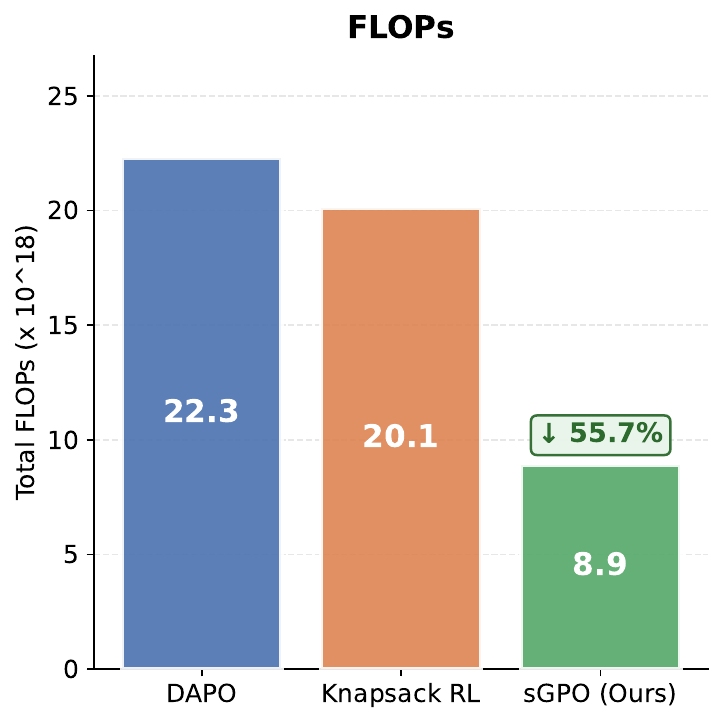}
        \caption{Total FLOPs ($\downarrow$).}
        \label{fig:sub1}
    \end{subfigure}
    \hfill
    \begin{subfigure}[b]{0.3\linewidth}
        \centering
        \includegraphics[width=\linewidth]{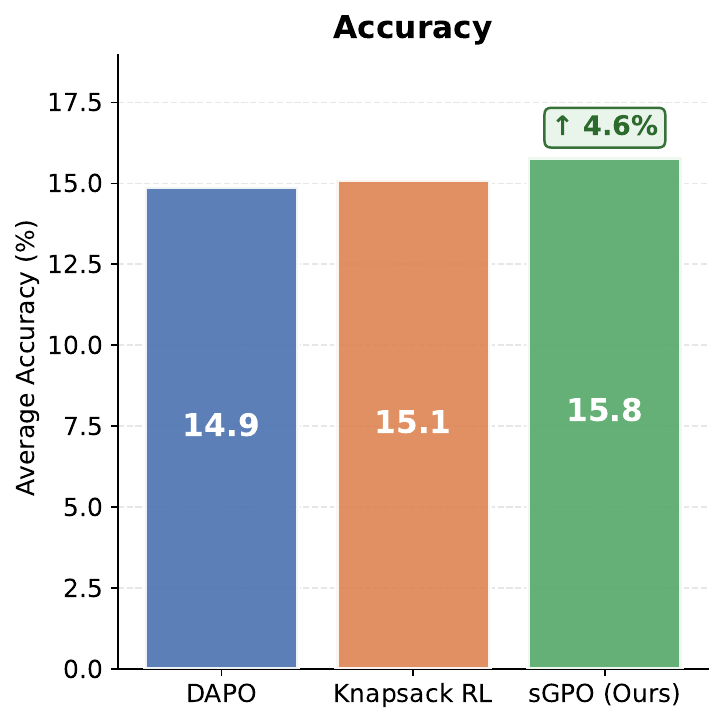}
        \caption{Average Accuracy ($\uparrow$).}
        \label{fig:sub2}
    \end{subfigure}
    \hfill
    \begin{subfigure}[b]{0.3\linewidth}
        \centering
        \includegraphics[width=\linewidth]{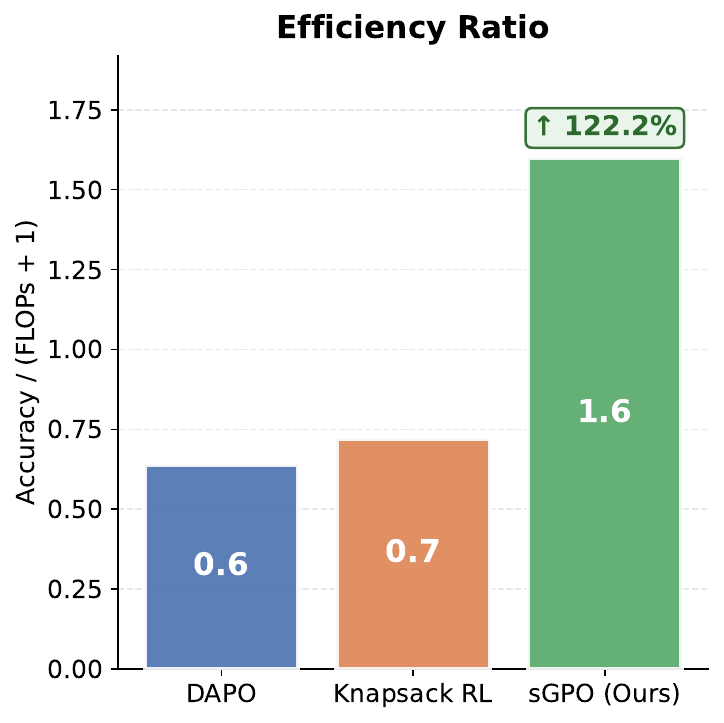}
        \caption{Efficiency Ratio ($\uparrow$).}
        \label{fig:sub3}
    \end{subfigure}
    \caption{Compute cost, accuracy, and efficiency of \method{} vs.\ baselines 
    on Qwen2.5-Math-7B. \emph{(a)}~\method{} requires 8.9~EF total - 2.5$\times$ 
    less than DAPO (22.3~EF) and 2.3$\times$ less than Knapsack RL (20.1~EF), 
    including a one-time profiling cost of 1.6~EF. \emph{(b)}~Despite this 
    reduction, \method{} achieves comparable average accuracy (15.8\%) across 5 math 
    benchmarks. \emph{(c)}~\method{} yields a 122.2\% better efficiency ratio, 
    confirming it extracts substantially more learning per unit of compute.}
    \label{fig:intro}
\end{figure*}

\paragraph{Compute Asymmetry}
This waste is particularly costly due to a fundamental \emph{compute asymmetry} 
in large language models: inference requires remarkably less FLOPs 
per token compared to policy training. Standard RLVR squanders these expensive 
training FLOPs on uninformative queries early on during training. 
The key insight is to exploit this asymmetry: spend a small upfront budget of \emph{cheap} inference FLOPs to profile the dataset, then drastically avoid wasting the consumption of \emph{expensive} training FLOPs.

Existing methods address parts of this problem; online filtering 
\citep{zheng2026act} removes uninformative queries but retains uniform $G$, saving only 2\% of FLOPs in our experiments; online solvers \citep{li2025knapsack} adapt $G$ per query but require a blind first epoch and achieve only 10\% savings. Both approaches operate online, recomputing decisions from training-time statistics at every step (Table~\ref{tab:comparison}).

We introduce \textit{sorted Group Policy Optimization} 
(\method{}), which optimizes the total computation of RLVR from a compute-allocation perspective. A single \emph{offline} profiling pass, costing ${\sim}7\%$ of the baseline compute, generates $N$ samples per query under the initial policy to obtain an empirical success rate $\hat{p}(q)$. This one quantity determines filtering thresholds, per-query group sizes ($\hat G \approx 1/\hat{p}$), and curriculum ordering, reducing total FLOPs by 60\% while maintaining accuracy and heavily boosting compute efficiency (Figure~\ref{fig:intro}).

\paragraph{Contributions} Our contributions are:
\begin{itemize}[noitemsep, topsep=2pt, leftmargin=*]
  \item[\emph{(i)}] We propose \method{}, a compute-efficient framework that uses 
  a single offline profiling pass to jointly filter data, allocate 
  adaptive group sizes, and construct an easy-to-hard curriculum.
  \item[\emph{(ii)}] We derive a sample-efficient heuristic for rollout group sizing in RLVR, showing that $\hat G(q) \approx 1/\hat{p}(q)$ provides the large non-zero advantage for a single successful rollout.
  \item[\emph{(iii)}] We empirically validate \method{} on math and science 
  reasoning benchmarks, where it matches or improves upon fixed-budget 
  DAPO while reducing training FLOPs by 60\% without loss of performance.
\end{itemize}

\begin{figure*}[t]
\centering
\resizebox{.95\textwidth}{!}{%

\begin{tikzpicture}[>=Stealth,
    every node/.style={font=\normalsize},
    stage/.style={rounded corners=8pt, draw=#1!60, fill=#1!2, inner sep=0pt, line width=1pt},
    stitle/.style={font=\large\bfseries, text=#1!80!black},
    response/.style={minimum width=0.4cm, minimum height=0.4cm, draw=slatelt, line width=0.5pt, inner sep=0pt},
    bigarr/.style={->, very thick, draw=slatelt, line width=2pt},
]

\node[stage=slate, minimum width=3.8cm, minimum height=4.5cm] (dataset) at (4.5, 0) {};
\begin{scope}[shift={(4.5,0)}]
    \node[font=\normalsize\bfseries, text=slate!80!black] at (0, 1.8) {Dataset};
    \foreach \i/\yoff in {1/0.9, 2/0.35, 3/-0.2, 4/-0.75} {
        \node[rounded corners=3pt, draw=slatelt, fill=white, minimum width=2.5cm, minimum height=0.4cm, font=\normalsize, text=slate] at (0, \yoff) {query $q_{\i}$};
    }
    \node[font=\normalsize, text=slatelt] at (0, -1.3) {$\vdots$};
\end{scope}

\node[stage=oisky, minimum width=7.0cm, minimum height=4.0cm] (profile) at (12.5, 0) {};
\begin{scope}[shift={(12.5,-0.5)}]
    \node[stitle=oiblue] at (0, 1.8) {(a) Profiling};

    \node[rounded corners=3pt, draw=oiblue!40, fill=bgsky, minimum width=1.3cm, minimum height=0.45cm, font=\normalsize\bfseries, text=oiblue] (pq) at (-2.0, 0.5) {$q_i$};
    \draw[->, thick, oiblue!40] (-1.3, 0.5) -- (-0.4, 0.5);

    \node[font=\normalsize, text=oiblue!60, above] at (2.3, 0.4) {\textbf{$N{=}8$}};
    \node[response, fill=correctfill] at (0.0, 0.8) {};
    \node[response, fill=wrongfill]   at (0.45, 0.8) {};
    \node[response, fill=correctfill] at (0.9, 0.8) {};
    \node[response, fill=wrongfill]   at (1.35, 0.8) {};
    \node[response, fill=wrongfill]   at (0.0, 0.35) {};
    \node[response, fill=correctfill] at (0.45, 0.35) {};
    \node[response, fill=wrongfill]   at (0.9, 0.35) {};
    \node[response, fill=wrongfill]   at (1.35, 0.35) {};

    \draw[->, thick, oiblue!40] (0.675, -0.1) -- (0.675, -0.45);
    \node[font=\normalsize\bfseries, text=oiblue] at (0.675, -0.8) {$\hat{p}(q_i) = 3/8$};
\end{scope}

\draw[bigarr] (dataset.east) -- (profile.west);

\node[rounded corners=6pt, draw=oiblue!70, fill=bgsky, line width=1.5pt,
      minimum width=2.5cm, minimum height=1.0cm,
      font=\large\bfseries, text=oiblue] (phub) at (9.5, -3.5) {$\hat{p}(q)$};

\draw[bigarr, oiblue!60] (profile.south) -- (12.5, -3.5) -- (phub.east);

\draw[bigarr, oiorange!60, rounded corners=6pt] (phub.south) -- (9.5, -4.5) -- (0, -4.5) -- (0, -5.0);
\draw[bigarr, oigreen!60] (phub.south) -- (9.5, -5.0);
\draw[bigarr, oipurple!60, rounded corners=6pt] (phub.south) -- (9.5, -4.5) -- (19.0, -4.5) -- (19.0, -5.0);

\node[stage=oiorange, minimum width=7.5cm, minimum height=4.5cm] (filter) at (0, -7.5) {};
\begin{scope}[shift={(0,-7.5)}]
    \node[stitle={oiorange!80!black}] at (0, 1.8) {(b) Data Selection};

    \begin{scope}[shift={(-1.5, -0.2)}]
        \draw[thick, slatelt] (0, 0) -- (0, 1.6);
        \draw[thick, slatelt] (0, 0) -- (3.0, 0);
        \node[font=\normalsize, text=slatelt] at (1.5, -0.25) {$\hat{p}$};

        \fill[deadfill] (0.0, 0) rectangle (0.35, 1.4);
        \draw[pattern=north east lines, pattern color=deadgray!50] (0.0, 0) rectangle (0.35, 1.4);

        \fill[oiblue!25] (0.4, 0) rectangle (0.75, 0.25);
        \fill[oiblue!25] (0.8, 0) rectangle (1.15, 0.4);
        \fill[oiblue!25] (1.2, 0) rectangle (1.55, 0.6);
        \fill[oiblue!25] (1.6, 0) rectangle (1.95, 0.75);
        \fill[oiblue!25] (2.0, 0) rectangle (2.35, 0.6);

        \fill[deadfill] (2.4, 0) rectangle (3.0, 0.35);
        \draw[pattern=north west lines, pattern color=deadgray!50] (2.4, 0) rectangle (3.0, 0.35);

        \node[font=\scriptsize, text=deadgray] at (0.175, -0.5) {$\hat{p}{=}0$};
        \node[font=\scriptsize, text=oiblue] at (1.375, -0.5) {learnable};
        \node[font=\scriptsize, text=deadgray] at (2.7, -0.5) {$\hat{p}{>}0.75$};
    \end{scope}

    \node[font=\normalsize, text=oiorange!70!black, align=center] at (0.0, -1.50) {filter if $\hat{p}{=}0$ or $\hat{p}{>}0.75$};
\end{scope}

\node[stage=oigreen, minimum width=7.5cm, minimum height=4.5cm] (assigng) at (9.5, -7.5) {};
\begin{scope}[shift={(9.5,-7.5)}]
    \node[stitle={oigreen!80!black}] at (0, 1.8) {(c) Adaptive $G$};

    \node[font=\normalsize\bfseries, text=oigreen!80!black, anchor=west] at (-2.5, 0.8) {Easy: $G{=}2$};
    \node[response, fill=correctfill] at (-0.2, 0.8) {};
    \node[response, fill=wrongfill]   at (0.25, 0.8) {};

    \node[font=\normalsize\bfseries, text=oigreen!80!black, anchor=west] at (-2.5, 0.0) {Med: $G{=}4$};
    \node[response, fill=correctfill] at (-0.2, 0.0) {};
    \node[response, fill=wrongfill]   at (0.25, 0.0) {};
    \node[response, fill=wrongfill]   at (0.7, 0.0) {};
    \node[response, fill=wrongfill]   at (1.15, 0.0) {};

    \node[font=\normalsize\bfseries, text=oigreen!80!black, anchor=west] at (-2.5, -0.85) {Hard: $G{=}8$};
    \node[response, fill=correctfill] at (-0.2, -0.55) {};
    \node[response, fill=wrongfill] at (0.25, -0.55) {};
    \node[response, fill=wrongfill] at (0.7, -0.55) {};
    \node[response, fill=wrongfill] at (1.15, -0.55) {};
    \node[response, fill=wrongfill] at (-0.2, -0.95) {};
    \node[response, fill=wrongfill] at (0.25, -0.95) {};
    \node[response, fill=wrongfill] at (0.7, -0.95) {};
    \node[response, fill=wrongfill] at (1.15, -0.95) {};

    \node[font=\normalsize, text=oigreen!60!black, align=center] at (0.0, -1.7) {$G^\star \approx 1/\hat{p}$};
\end{scope}

\node[stage=oipurple, minimum width=7.5cm, minimum height=4.5cm] (train) at (19.0, -7.5) {};
\begin{scope}[shift={(19.0,-7.5)}]
    \node[stitle={oipurple!80!black}] at (0, 1.8) {(d) Curriculum};

    \node[rounded corners=4pt, draw=oipurple!40, fill=bgpurple, minimum width=1.4cm, minimum height=2.4cm, font=\normalsize\bfseries, text=oipurple!60!black, align=center] (model) at (1.6, 0.0) {$p_\theta$};

    \draw[->, thick, oipurple!60] (-1.6, 0.8) -- (model.west |- 0,0.8) node[midway, above=-1pt, font=\normalsize\bfseries, text=oipurple!70!black] {P1: $G{=}2$};
    \draw[->, thick, oipurple!60] (-1.6, 0.0) -- (model.west) node[midway, above=-1pt, font=\normalsize\bfseries, text=oipurple!70!black] {P2: $G{=}4$};
    \draw[->, thick, oipurple!60] (-1.6, -0.8) -- (model.west |- 0,-0.8) node[midway, above=-1pt, font=\normalsize\bfseries, text=oipurple!70!black] {P3: $G{=}8$};

    \draw[decorate, decoration={brace, amplitude=4pt}, oipurple!50, thick] (-1.8, 1.0) -- (-1.8, -1.0) node[midway, xshift=-14pt, font=\normalsize\bfseries, text=oipurple!80!black, rotate=90, anchor=center] {easy$\to$hard};

    \node[font=\normalsize, text=oipurple!60!black, align=center] at (0, -1.7) {sort by $\hat{p}$};
\end{scope}

\end{tikzpicture}
}
\caption{The \method{} pipeline. \textbf{(a) Profiling:} $N{=}8$ samples per query yield an empirical success rate $\hat{p}(q)$. This single signal drives all downstream decisions: \textbf{(b)} filtering queries with $\hat{p}{=}0$ or $\hat{p}{>}0.75$, \textbf{(c)} assigning rollout group sizes $\hat G \approx 1/\hat{p}$, and \textbf{(d)} ordering the curriculum from easy to hard by $\hat{p}$.}
\label{fig:pipeline}
\end{figure*}
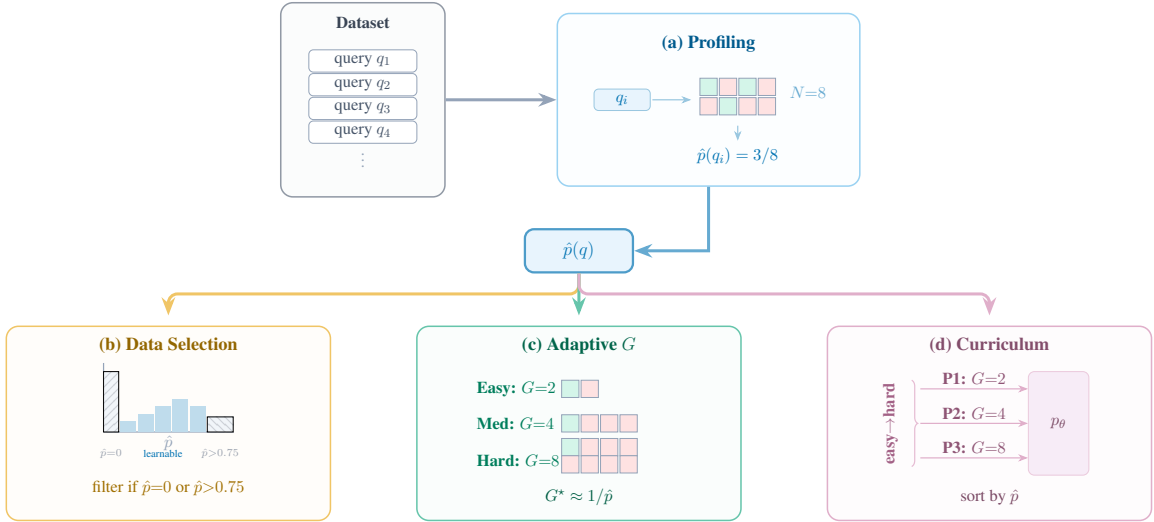

\section{Background}
\label{sec:background}

\paragraph{Policy Optimization}
GRPO-like~\citep{guo2025deepseek} algorithms sample a group of \(G\) rollouts per task from the model, \(\{\tau_i\}^G_{i=1} \sim p_{\theta}(\tau \mid q)\), and normalize their rewards into advantages:
\begin{equation}
    a_i = \frac{r_i - \bar{r}}{\sigma_r}, \quad i = 1, \ldots, G,
\end{equation}
where $\bar r = \frac{1}{G} \sum^G_{j=1} r_j$ and $\sigma^2_r = \frac{1}{G} \sum^G_{j=1}(r_j - \bar r)^2$ are sample statistics for a finite group $G$.

More generally, GRPO-like objectives may also include additional components, such as divergence regularization with respect to a reference model, clipping, importance weighting for batched training, and alternative definitions of the advantage~\citep{guo2025deepseek, liu2025understanding,yu2025dapo}. 
In this work, however, we focus on the core ingredients of the method---policy optimization and advantage computation---because our approach depends only on statistics that can be computed for any GRPO-like algorithm, as long as a rollout group of size \(G\) is sampled.

Each token in rollouts \(i\) receives the same advantage \(a_i\), and the policy is updated via a clipped surrogate objective. This approach works well when at least some rollouts succeed. However, when all \(G\) rollouts receive the same reward, every advantage is zero and the gradient vanishes---leaving the policy with no learning signal.

Given a dataset $\mathcal{D} = \{q_j\}_{j=1}^{|\mathcal{D}|}$ and a fixed group size $G$, the batch estimator~\citep{williams1991function} is:
\begin{equation}
\mathcal{\hat F}(\theta) = \frac{1}{|\mathcal{D}|\,G} \sum^{|\mathcal{D}|}_{j=1}  \sum^{G}_{i=1} a(\tau_i, q_j),
\label{eq:pg-batch}
\end{equation}

and corresponding empirical gradients $\nabla_{\theta} \mathcal{\hat F}(\theta)$:
\begin{equation}
 \frac{1}{|\mathcal{D}|~G} \sum^{|\mathcal{D}|}_{j=1}  \sum^{G}_{i=1} a(\tau_i, q_j) \nabla_{\theta} \log p_{\theta}(\tau_i \mid q_j).
\label{eq:pg-batch-gradients}
\end{equation}

These design choices arise because the sampler lacks a 
model-aware estimate of query difficulty---a gap that \method{} fills.

\paragraph{Inference-Time Scaling}
Inference-time scaling (ITS,~\citep{brown2024large,snell2024scaling}) improves reasoning by allocating additional compute at generation time, typically through multiple sampled candidates and verifier-based selection~\citep{lightman2023let}. In tasks with verifiable rewards, ITS can expose latent capability without changing model parameters. Following such properties, we use a cheap ITS pass before training to estimate how often the initial policy solves each query.

\section{Related Work}
\label{sec:related}

\begin{table}[ht!]
\centering
\resizebox{.9\columnwidth}{!}{%
\renewcommand{\arraystretch}{1.2}
\begin{tabular}{@{}lcccc@{}}
\toprule
\multirow{2}{*}{Method} & Offline & Data & Adaptive & Curriculum \\
 & Profiling & Selection & Allocation & Building \\
\midrule
GRESO       & \xmark          & \cmark       & \xmark     & \xmark \\
Knapsack RL & \xmark          & \xmark       & \cmark     & \xmark \\
\rowcolor[HTML]{D9EAD3} \method{} (Ours) & \textbf{\cmark} & \textbf{\cmark} & \textbf{\cmark} & \textbf{\cmark} \\
\bottomrule
\end{tabular}%
}
\caption{Comparison of compute allocation strategies for RLVR training. GRESO uses online profiling, filter-based data selection, uniform allocation, and random ordering; Knapsack RL uses online profiling, no data selection, solver-based allocation, and random ordering; \method{} uses offline profiling, filter + mix data selection, inference-based allocation, and easy-to-hard curriculum building. Only \method{} combines all four components through a cheap offline ITS profiling pass.}
\label{tab:comparison}
\end{table}

\paragraph{Adaptive Rollout Allocation}
Methods for non-uniform compute allocation in RLVR fall into two 
families. The first filters queries: GRESO~\citep{zheng2026act} skips 
prompts with consistent reward history, and DEPO~\citep{zhao2026difficulty} 
combines PageRank-based selection with DPP diversity sampling. The 
second varies $G$ per query using online difficulty estimates: 
Knapsack~RL~\citep{li2025knapsack} via knapsack optimization, 
VIP~\citep{nguyen2026adaptive} and CoBA-RL~\citep{yao2026coba} via 
probabilistic models, AR3PO~\citep{zhang2025improving} via Bayesian 
and replay-based approaches, and GDRO~\citep{panaganti2026group} via 
adversarial DRO under a fixed mean-budget constraint. All operate 
\emph{online}, recomputing decisions from training-time statistics at 
every step or epoch, which is computationally suboptimal. \method{} instead profiles once offline: a single 
$N{=}8$ inference pass jointly determines filtering, group sizes, and 
curriculum ordering before training begins (Table~\ref{tab:comparison}).

\paragraph{Curriculum Learning for Reasoning}
Curriculum learning~\citep{bengio2009curriculum} proposes scheduling 
data from easy to hard. In RLVR, E2H~Reasoner~\citep{parashar2025curriculum} 
provides convergence bounds for easy-to-hard ordering, 
SEC~\citep{chen2025self} constructs self-evolving curricula from 
advantage proxies, VCRL~\citep{jiang2025vcrl} uses reward variance, 
TACLer~\citep{lai2026tacler} and ADCL~\citep{zhang2025learning} 
re-estimate difficulty during training, and Light-R1~\citep{wen2025light} 
stages curriculum across SFT, DPO, and RL. All treat ordering as an 
independent design choice; in \method{}, the curriculum is a direct 
consequence of $\hat{p}(q)$---the same signal that drives filtering 
and group sizing.

\paragraph{Compute-optimal Scaling}
\citet{snell2024scaling} show that adaptive inference-time compute 
outperforms uniform scaling, and \citet{damani2024learning} extend 
this to input-adaptive computation. \method{} applies this perspective 
at \emph{training} time, using the profiling pass to concentrate 
expensive training FLOPs on queries where the learning signal is highest.

\section{Method}
\label{sec:method}

\method{} optimizes the compute allocation of RLVR training from a sample efficiency perspective: rather than assigning a fixed rollout budget to every query, a small upfront profiling pass estimates $\hat{p}(q)$, which then guides filtering, rollout allocation, and curriculum design from a single offline signal (Figure~\ref{fig:pipeline}).

\subsection{Profiling via Inference-Time Scaling}

The foundation of \method{} is a single, offline profiling pass that measures the empirical success rate of the initial policy $p_\theta$ on every query in the dataset. For each query $q$, we generate $N$ parallel samples and compute the empirical success rate $\hat{p}(q)$:
\begin{equation}
    \hat{p}(q) = \frac{n_{\text{profiling}}(q)}{N},
\end{equation}

where $n_{\text{profiling}}(q)$ is the number of correct responses among the $N$ profiling samples.

Rather than treating difficulty as a fixed extrinsic property, this profiling defines difficulty relative to the base policy: a query is only as hard as how often $p_\theta$ fails on it.

The profiling pass is cheap: at $N = 8$ samples per query, profiling the full dataset costs approximately one GPU-hour---a one-time cost amortized across all subsequent training phases.

\subsection{Sample-Efficient Advantage}
\label{subsec:allocation}

For binary rewards\footnote{We assume a Bernoulli model over the sample rewards, i.e., $r \sim \mathcal{B}(p)$.} and finite group $G$, the realized advantage $a_i$ of
rollout $i$ with $n = \sum_{j=1}^{G} r_j$ successes in the group is
\begin{equation}
  a_i = \frac{r_i - \bar{r}}{\sigma_r}, \quad
  \bar{r} = \frac{1}{G}\sum_{j=1}^{G} r_j, \quad
  \sigma_r = \sqrt{\bar{r}(1-\bar{r})}.
  \label{eq:realized_advantage}
\end{equation}
For a successful rollout ($r_i = 1$), this simplifies to
\begin{equation}
  a_i\big|_{r_i=1} 
  = \dfrac{1 - n/G}{\sqrt{n/G~(1 - n/G)}}
  = \sqrt{\frac{G-n}{n}},
  \label{eq:realized_success}
\end{equation}
which vanishes when $n = G$ (all rollouts correct)---confirming the
zero-gradient collapse that the filtering step targets. Conversely, if $n = 0$, there are no successful rollouts to carry the advantage, again yielding no useful signal.

Between these extremes, the advantage of a successful rollout $a_i|_{r_i=1} = \sqrt{(G-n)/n}$ strictly decreases as the number of successes $n$ increases. \emph{Therefore, to maximize sample efficiency, we want the smallest group that still surfaces a success}. Targeting exactly $n=1$ achieves this, yielding the maximum non-zero advantage of $\sqrt{G-1}$ for a successful rollout.

\paragraph{Learning Signal vs Efficiency}
While a balanced split of successes and failures ($n = G/2$) can maximize the intra-group reward variance $\sigma_r^2$ (Eq.~\ref{eq:realized_advantage}) to provide strong gradient update in terms of global learning signal, achieving this balance on hard queries ($p \ll 1$) requires prohibitively large group sizes. Instead, targeting $n=1$ explicitly prioritizes compute efficiency by extracting the maximum advantage from a single successful rollout. The goal of \method{} is to \emph{maximize learning per FLOP}: it targets the smallest $G$ that reliably produces a non-zero gradient, rather than the $G$ that maximizes the total learning signal.

Under a Binomial model, the expected number of successes in a group of size $G$ is $\mathbb{E}[n] = G p$, where $p$ is the true underlying success probability of the policy for task $q$. By setting this expectation to our target of a single success ($\mathbb{E}[n] = 1$), we obtain the ideal theoretical group size, under the base policy, for sample efficiency: $G = \frac{1}{p}$.

Since the true probability $p$ is unknown, we substitute it with the empirical estimate $\hat{p}(q)$ obtained from the offline profiling pass ($N=8$). This motivates the approximate training group size heuristic:
\begin{equation}
    \hat G(q) \approx \frac{1}{\hat{p}(q)}, \quad
    \hat{p}(q) = \frac{n_{\mathrm{profiling}}(q)}{N},
    \label{eq:optimal_G}
\end{equation}

where $\hat{p}(q)$ is estimated from an offline profiling pass with budget $N$, and $\hat{G}(q)$ denotes the resulting approximate per-query training allocation.
Because $N$ is fixed during profiling whereas $G$ is chosen per query during training, the relation $\hat{G}(q) = 1/\hat{p}(q)$ should be understood as a practical heuristic.

Choosing $G \approx 1/p$ balances the two failure modes identified above: too few rollouts risk no successes, too many produce redundant ones.

\begin{figure}[t]
\centering
\includegraphics[width=.9\columnwidth]{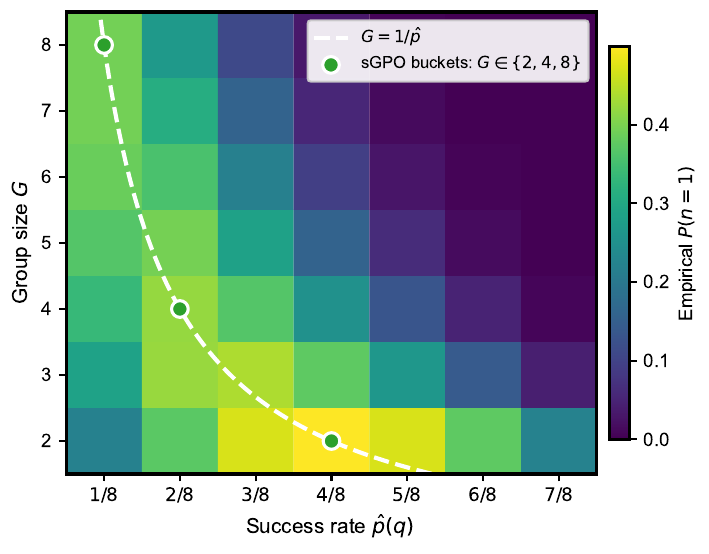}
\caption{Empirical single-success probability $P(n{=}1)$ as a function of $\hat{p}(q)$ and $G$, computed by subsampling the profiling data. The bright ridge tracks the theoretical frontier $G = 1/\hat{p}$ (dashed), where exactly one success per group is most likely. \method{}'s discrete assignments $G \in \{2,4,8\}$ (green dots) follow this ridge. Above the frontier, groups contain redundant successes; below it, groups risk no correct rollout.}
\label{fig:g_landscape}
\end{figure}

\subsection{Training Strategies from $\hat{p}$}
\label{subsec:decisions}

The profiled success rate $\hat{p}(q)$ determines three training decisions, each derived directly from the same quantity: which queries to train on, how many rollouts each receives, and in what order they are presented.

\subsubsection{Data Selection}
\label{subsubsec:selection}

Using the profiled success rates $\hat{p}_j$ and a filtering threshold $t$, we partition the training dataset $\mathcal{D}$ into three subsets:
\begin{align}
\begin{split}
    \mathcal{D}_{\mathrm{trivial}}      &= \{q_j \in \mathcal{D} \mid \hat{p}_j > t\}, \\
    \mathcal{D}_{\mathrm{unsolved}}  &= \{q_j \in \mathcal{D} \mid \hat{p}_j = 0\},  \\
    \mathcal{D}_{\mathrm{learnable}} &= \{q_j \in \mathcal{D} \mid 0 < \hat{p}_j \leq t \}.
\end{split}
\label{eq:data-learnable}
\end{align}

\paragraph{Trivial queries} ($\hat{p}_j > 0.75$) are removed entirely. By the analysis in Section~\ref{subsec:allocation}, they produce near-zero advantage and consume training FLOPs without benefit.

\paragraph{Unsolved queries} ($\hat{p}_j = 0$ at $N = 8$) produce no learning signal in expectation under the current policy and is therefore excluded from the primary training clusters. However, discarding them entirely would forgo long-horizon exploration. Instead, we subsample a fraction $\alpha$ of $\mathcal{D}_{\mathrm{unsolved}}$ and mix it into every training phase as $\tilde{\mathcal{U}}_g \subset \mathcal{D}_{\mathrm{unsolved}}$, \emph{encouraging the policy to continue searching over high-complexity reasoning paths throughout training}. We use $\alpha = 10\%$ and ablate this choice in Section~\ref{sec:ablation}. The training cluster for phase $g$, augmented with the unsolved subsample, is:
\begin{equation}
    \bar{C}_g = C_g \cup \tilde{\mathcal{U}}_g, \quad g \in \{2, 4, 8\}.
    \label{eq:augmented_cluster}
\end{equation}

\subsubsection{Adaptive Group Size}
\label{subsubsec:groupsize}

Learnable queries ($0 < \hat{p}_j \leq t$) form the core training set. Since training requires a fixed group size per batch, we discretize $\hat G(q) \approx 1/\hat{p}(q)$ into three power-of-two buckets:
\begin{equation}
b(q) = \begin{cases}
  2 & \text{if } \hat{p}(q) \in (1/4,\, t], \\
  4 & \text{if } \hat{p}(q) \in (1/8,\, 1/4], \\
  8 & \text{if } \hat{p}(q) \in (0,\, 1/8],
\end{cases}
\quad G_j = b(q_j),
\label{eq:bucket_map}
\end{equation}
where $t = 0.75$ is the trivial-query threshold. With profiling budget $N{=}8$, the success rate $\hat{p}(q)$ takes discrete values in $\{0, 1/8, \ldots, 1\}$, so the bucket boundaries align with the profiling resolution. At $\hat{p} \in \{1/8, 1/4, 1/2\}$, the heuristic $\hat G = 1/\hat{p}$ coincides exactly with the assigned $G$ and $\mathbb{E}[n] = 1$. At other values (e.g., $\hat{p} = 3/8$ with $G{=}2$), $\mathbb{E}[n] = 0.75$, which remains in the region where gradient signal is non-zero (Figure~\ref{fig:g_landscape}).

Applying the bucket map partitions $\mathcal{D}_{\mathrm{learnable}}$ into three rollout clusters:
\begin{equation}
    C_g = \{q_j \in \mathcal{D}_{\mathrm{learnable}} \mid G_j = g\},
    \label{eq:clusters}
\end{equation}
for $g \in \{2, 4, 8\}$.

\subsubsection{Curriculum Ordering}
\label{subsubsec:curriculum}

The three clusters are traversed in ascending difficulty order:
\begin{equation}
    \theta^\star = \mathrm{SeqTrain}\!\left(
        \hat{\mathcal{F}}_2 \;\to\; \hat{\mathcal{F}}_4 \;\to\; \hat{\mathcal{F}}_8
    \right).
    \label{eq:seq_train}
\end{equation}
Starting with $G = 2$ clusters (high-$\hat{p}$ queries) ensures the policy first refines capabilities where the advantage signal from Eq.~\eqref{eq:realized_success} is reliable and dense, before progressively encountering harder queries that require larger groups to surface successful trajectories. The consistent injection of unsolved queries across all phases prevents the policy from overfitting to easy data in early phases.

\subsection{Training Objective}
\label{subsec:objective}

For each phase $g$, we optimize the cluster-specific advantage objective:
\begin{equation}
    \hat{\mathcal{F}}_g(\theta) = \frac{1}{|\bar{C}_g|}
    \sum_{q_j \in \bar{C}_g} \frac{1}{G_j}
    \sum_{i=1}^{G_j} a(\tau_i, q_j),
    \label{eq:cluster_objective-sgpo}
\end{equation}
with $G_j = g$ for all $q_j \in \bar{C}_g$. When $G_j = G$ for all $j$ and no partitioning or mixing is applied, this reduces exactly to the standard fixed-budget objective in Eq.~\eqref{eq:pg-batch}.

\paragraph{Gradients}
The gradients for the \method{} estimator $\hat{\mathcal{F}}_g(\theta)$ for group cluster $g$ are:
\begin{equation}
\frac{1}{|\bar{C}_g|}
    \sum_{q_j \in \bar{C}_g} \frac{1}{G_j}
    \sum_{i=1}^{G_j} a(\tau_i, q_j) \nabla_{\theta} \log p_{\theta}(\tau_i \mid q_j).
\label{eq:sgpo-gradients}
\end{equation}
By construction, $a(\tau_i, q_j)$ provides an effective learning signal over each group query $G_j$ and group cluster $\tilde C_g$ in expectation.

\paragraph{Implementation}

We instantiate \method{} on top of DAPO~\citep{yu2025dapo}. The full pipeline is summarized in Algorithm~\ref{alg:sorted_g}.

\section{Experiments}
\label{sec:experiment}

We evaluate \method{} on mathematical reasoning and scientific question answering. Our experiments ask whether offline profiling improves accuracy per FLOP over online and uniform baselines, and whether these gains generalize across domains with different structures.

\paragraph{Models}
We conduct our experiments with both pretrained and instruction-tuned models. We use Qwen2.5-Math (base) series ~\citep{yang2024qwen2} as our primary models for mathematical reasoning experiments and Qwen3-4B-Instruct-2507~\citep{yang2025qwen3} for cross-domain science experiments.

\paragraph{Datasets}
For mathematics, we train on DAPO-Math-17k~\citep{yu2025dapo}, a curated dataset of 14,116 English mathematical reasoning problems with verifiable answers. For science reasoning, we use SciKnowEval~\citep{feng2024sciknoweval}, which contains undergraduate-level scientific multiple-choice questions across four domains: chemistry, physics, biology, and materials science.

\paragraph{Baselines}
We compare \method{} against three baselines (Table~\ref{tab:comparison}). DAPO~\citep{yu2025dapo} uses a uniform rollout group size \(G{=}8\) with dynamic sampling, filtering zero-variance groups only after rollout generation and ordering data randomly. GRESO~\citep{zheng2026act} performs filtering before rollout by probabilistically skipping prompts with zero-variance history, but retains uniform \(G\) and random ordering. Knapsack RL~\citep{li2025knapsack} allocates \(G\) online via knapsack optimization over per-query success rates from training history, but without pre-training profiling its first epoch is identical to uniform allocation.

\paragraph{Evaluation}
We evaluate on mathematical reasoning (AIME 2024/2025/2026 and HMMT 
February 2025/2026~\citep{balunovic2025matharena}) and scientific 
reasoning (SciKnowEval-L3~\citep{feng2024sciknoweval} 
across chemistry, physics, biology, and materials science). 
Throughout, we report avg@16 accuracy; full setup is in 
Appendix~\ref{appx:details}.

\subsection{Main Results}

\begin{table*}[ht!]
\centering
\resizebox{.9\textwidth}{!}{ 
\begin{tabular}{lccc | cccccc}
\toprule
 & \multicolumn{3}{c}{\textbf{Efficiency (FLOPs) $\downarrow$}} & \multicolumn{6}{c}{\textbf{Performance (Accuracy) $\uparrow$}} \\
\cmidrule(lr){2-4} \cmidrule(l){5-10}
{Method} & {Profiling} & {Training} & {Total} & {AIME24} & {AIME25} & {AIME26} & {HMMT25} & {HMMT26} & {Avg} \\
\midrule
\\
\multicolumn{10}{l}{\emph{Qwen2.5-Math-1.5B}} \\
\midrule
Base model & --- & --- & --- & 9.4 & 4.0 & 6.7 & 0.6 & 1.5 & 4.4 \\
GRESO & 0.0 & 4.3 & 4.3 & 11.5 & 5.8 & 8.5 & 0.2 & \textbf{3.4} & 5.9\\
DAPO & 0.0 & 4.4 & 4.4 & 15.0 & 4.6 & 8.1 & 0.8 & 1.7 & 6.0 \\
Knapsack RL & 0.0 & 4.7 & 4.7 & 16.5 & \textbf{6.8} & \textbf{9.0} & 0.8 & 3.0 & \textbf{7.2}  \\
\rowcolor[HTML]{D9EAD3} \method{} (ours) & 0.4 & \textbf{1.1} & \textbf{1.5} & \textbf{16.7} & 5.8 & 8.8 & \textbf{1.0} & 2.7 & 7.0 \\
\midrule
\\
\multicolumn{10}{l}{\emph{Qwen2.5-Math-7B}} \\
\midrule
Base model & --- & --- & --- & 16.9 & 5.4 & 7.1 & 1.0 & 2.7 & 6.6 \\
GRESO & 0.0 & 21.9 & 21.9 & 26.7 & 12.9 & 11.5 & 2.1 & 6.1 & 11.9\\
DAPO & 0.0 & 22.3 & 22.3 & 29.8 & 12.5 & 14.0 & 2.9 & 5.4 & 14.9 \\
Knapsack RL & 0.0 & 20.1 & 20.1 & 28.9 & \textbf{14.8} & 11.9 & \textbf{3.8} & 6.1 & 15.1 \\
\rowcolor[HTML]{D9EAD3} \method{} (ours) & 1.6 & \textbf{7.3} & \textbf{8.9} & \textbf{31.0} & 13.5 & \textbf{15.4} & 2.5 & \textbf{6.6} & \textbf{15.8} \\
\bottomrule
\end{tabular}
}
\caption{Main results on mathematical reasoning benchmarks with different model scales. FLOPs are reported in ExaFLOPs ($\times 10^{18}$); evaluation scores are avg@16 accuracy (\%). Profiling FLOPs denotes the one-time inference cost; Train FLOPs denotes the RL training loop cost. \method{} matches or exceeds the strongest baseline across both model scales while requiring 2.5--3.1$\times$ fewer total FLOPs than the strongest baseline, even after accounting for the upfront profiling budget.}
\label{tab:main_results}
\end{table*}

\paragraph{Mathematical Reasoning}
Table~\ref{tab:main_results} shows that \method{} achieves the best 
compute-performance trade-off across both model scales. At 7B, it 
obtains the highest average accuracy (15.8\%) while requiring only 
8.9~EF total - 2.5$\times$ less than DAPO and 2.3$\times$ less than 
Knapsack~RL. The one-time profiling cost of 1.6~EF reduces the RL 
training budget to 7.3~EF, well below DAPO (22.3~EF) and Knapsack~RL 
(20.1~EF), and \method{} achieves the best individual scores on 
AIME~2024, AIME~2026, and HMMT~2026.

At 1.5B, \method{} again leads on AIME~2024 (16.7\%) at just 1.5~EF 
total - roughly 3$\times$ less than DAPO (4.4~EF) and Knapsack~RL 
(4.7~EF) - with a profiling overhead of only 0.4~EF. GRESO provides 
negligible savings and weaker accuracy at both scales; Knapsack~RL 
matches \method{} in accuracy but captures only modest FLOPs 
reductions. Figure~\ref{fig:flop_efficiency} confirms the trend: 
\method{} maintains a superior accuracy-compute frontier throughout 
training and reaches comparable peak performance substantially earlier.

\paragraph{Scientific Reasoning}

\begin{table*}[t]
\centering
\resizebox{.8\linewidth}{!}{%
\begin{tabular}{lc| cccccc}
\toprule
& \multicolumn{1}{c}{\textbf{Efficiency (FLOPs) $\downarrow$}} & \multicolumn{5}{c}{\textbf{Performance (Accuracy) $\uparrow$}} \\
\cmidrule(lr){2-2} \cmidrule(l){3-8} 
{Method} & Total (Profiling) & {Chemistry} & {Physics} & {Biology} & {Materials} & {Avg} & {w Avg}\\
\midrule
Base model & --- & 36.8 & 53.0 & 25.5 & 64.9 & 45.1 & 44.2 \\
DAPO & 5.2 (0.0) & 63.3 & \textbf{66.6} & 26.8 & \textbf{74.8} & 57.9 & 61.6 \\
\rowcolor[HTML]{D9EAD3} \method{} (ours) & \textbf{1.9 (0.4)} & \textbf{64.3} & 65.7 & \textbf{27.8} & 74.0 & \textbf{58.0} & \textbf{61.9}\\
\bottomrule
\end{tabular}%
}
\caption{Cross-domain results on SciKnowEval using a generalist model, Qwen3-4B-Instruct. FLOPs are total ExaFLOPs (\( \times 10^{18} \)), including \method{}'s one-time profiling cost of 0.4~EF; scores are avg@16 accuracy (\%). \method{} matches DAPO at 2.7\( \times \) fewer FLOPs.}
\label{tab:science_results}
\end{table*}

To evaluate \emph{cross-domain generalization}, we apply \method{} to SciKnowEval using Qwen3-4B-Instruct (Table~\ref{tab:science_results}). \method{} requires only 1.9~EF total FLOPs, compared with 5.2~EF for DAPO, yielding a \(2.7\times\) reduction in compute. Despite this substantially lower budget, \method{} slightly improves over DAPO in both average accuracy and weighted average accuracy. The one-time profiling cost is just 0.4~EF, accounting for 21\% of the total budget. \method{} also achieves the best results in Chemistry, Biology, average accuracy, and weighted average accuracy, while remaining competitive in Physics and Materials. These results suggest that the efficiency gains from \method{} extend beyond mathematical reasoning to scientific reasoning tasks with different domain structure and reward characteristics.

\paragraph{Key Takeaways}
Across both domains, \method{} delivers the best compute-performance 
trade-off: 2.5--3.1$\times$ fewer total FLOPs than DAPO on mathematics 
and 2.7$\times$ fewer on science, while matching or slightly exceeding 
accuracy in both settings. The consistent gains confirm that offline 
profiling generalizes beyond mathematical reasoning - the same 
cheap inference pass that drives efficiency on AIME and HMMT 
transfers directly to multi-domain scientific QA.

\subsection{Analysis}
\label{sec:analysis} 

\paragraph{Inference-training Cost Asymmetry}
At 7B scale, 1.6~EF of inference-only profiling (${\sim}7\%$ of DAPO's total budget) determines which queries to keep, what group size each needs, and in what order to present them. Because every training step uses these decisions, the model learns efficiently from step one: no wasted rollouts on trivial or unsolvable queries, no oversized groups on easy problems. The result is a training reduction from 22.3~EF to 7.3~EF, with the 1.6~EF profiling cost paid in cheap inference tokens (Figure~\ref{fig:sub3}).

\paragraph{Online Baselines}
GRESO and Knapsack~RL both estimate difficulty online from training-time rewards. GRESO filters queries but retains uniform $G{=}8$ for all kept queries, saving 2\% of FLOPs (21.9 vs.\ 22.3~EF at 7B) while degrading AIME~2024 accuracy from 29.8\% to 26.7\%. Filtering alone does not address per-query compute waste: an easy query at $\hat{p} = 0.5$ still generates $\mathbb{E}[n] = 4$ redundant successes per group. Knapsack~RL adapts $G$ per query but requires a blind first epoch at uniform $G{=}8$ before its solver has signal, limiting total savings to 10\% (20.1 vs.\ 22.3~EF).

\paragraph{Efficiency Decomposition}
\method{}'s 15.0~EF training reduction over DAPO at 7B decomposes into two sources. Filtering (removing trivial and unsolved queries) reduces the training set from 14,116 to 7,525 queries, accounting for 10.4~EF (69\% of the savings). Adaptive group sizing reduces the average $G$ from 8 to 4.1 across the remaining queries, saving an additional 4.6~EF (31\%). Neither component alone matches \method{}'s total savings: filtering without adaptive $G$ still wastes rollouts on easy queries (as GRESO demonstrates), while adaptive $G$ without filtering still trains on unsolvable and trivial problems.

\subsection{Ablations}
\label{sec:ablation}

\begin{table}[ht!]
\centering
\small
\setlength{\tabcolsep}{3pt}
\resizebox{.9\linewidth}{!}{%
\begin{tabular}{lcc}
\toprule
\textbf{Method} & \textbf{FLOPs} ($\times 10^{18}$) & \textbf{AIME24}$_{\text{avg@16}}$ \\
\midrule
\method{} & \textbf{8.9} & \textbf{31.0} \\
~$-$ Curriculum & 8.9 & 29.6 \\
~$-$ Adaptive $G$ & 10.7 & 27.5 \\
~$-$ Filtering (DAPO) & 22.3 & 29.8 \\
\bottomrule
\end{tabular}
}
\caption{Component ablation of \method{} on Qwen2.5-Math-7B. Rows are cumulative: each removes one additional component from the full method, ending at DAPO.}
\label{tab:ablation}
\end{table}

Table~\ref{tab:ablation} cumulatively removes \method{} components.

\paragraph{Curriculum} 
Removing easy-to-hard ordering drops accuracy from 31.0\% to 29.6\% with no change in FLOPs. Ordering queries by difficulty costs nothing in compute but improves sample efficiency by 1.4pp: the model builds capability on high-$\hat{p}$ queries in the $G{=}2$ phase before encountering low-$\hat{p}$ queries in the $G{=}8$ phase.

\paragraph{Adaptive Group} 
Additionally removing per-query group sizes (reverting to uniform $G{=}8$) increases FLOPs from 8.9 to 10.7~EF and drops accuracy to 27.5\%. Without adaptive sizing, every kept query generates eight rollouts regardless of difficulty, wasting compute on queries where fewer rollouts would suffice.

\paragraph{Filtering} 
Filtering without adaptive $G$ (27.5\%) performs worse than no filtering at all (DAPO, 29.8\%). Removing queries shrinks the dataset but does not reduce per-query cost, so the model sees fewer problems with the same per-problem waste. Filtering is beneficial only when adaptive $G$ reduces the cost of the queries that remain (consistent with GRESO; Section~\ref{sec:analysis}).

\begin{table}[t]
\centering
\small
\setlength{\tabcolsep}{3pt}
\begin{tabular}{lcc}
\toprule
\textbf{Mixing $\alpha$} & \textbf{FLOPs} ($\times 10^{18}$) & \textbf{AIME24}$_{\text{avg@16}}$ \\
\midrule
$0\%$ & 5.9 & 28.3 \\
$10\%$ & 8.9 & \textbf{31.0} \\
$25\%$ & 9.4 & 29.8 \\
$30\%$ & 10.2 & 29.2 \\
\bottomrule
\end{tabular}
\caption{Effect of unsolved mixing ratio $\alpha$ on Qwen2.5-Math-7B. $\alpha$ controls the fraction of $\mathcal{D}_{\mathrm{unsolved}}$ mixed into each curriculum phase.}
\label{tab:ablation_mixing}
\end{table}

\paragraph{Unsolved Mixing Ratio} Table~\ref{tab:ablation_mixing} varies $\alpha$, the fraction of $\mathcal{D}_{\mathrm{unsolved}}$ mixed into each curriculum phase. Accuracy peaks at $\alpha{=}10\%$ (31.0\%). Lower values drop accuracy (e.g., 28.3\% at $\alpha{=}0\%$) by starving the policy of hard exploration targets, though $\alpha{=}0\%$ remains viable for extreme compute constraints (5.9~EF). Higher values degrade performance by diluting the learnable set with zero-gradient rollouts.

\paragraph{Multiple Epochs}
All main results use two epochs with stale profiling assignments (same $\hat G$ as epoch~1). As detailed in Appendix~\ref{appx:multi_epoch}, re-profiling the dataset after epoch~1 degrades accuracy (${\sim}25\%$ vs.\ ${\sim}32\%$ for stale replay) because it over-concentrates training on overly hard, low-signal problems. Retaining the original assignments preserves necessary difficulty diversity.

\section{Conclusion}
\label{sec:conclusion}
Standard RLVR training allocates compute uniformly, wasting FLOPs on queries the policy has already mastered or cannot yet solve. By using a single offline profiling pass to drive data filtering, group sizing, and curriculum ordering, \method{} trades cheap inference FLOPs for expensive training FLOPs to match baseline accuracy while reducing total compute by 2.5-3.1$\times$. More broadly, these results suggest that a small investment in offline difficulty estimation can substitute for a large portion of online compute in RLVR.

\clearpage

\paragraph*{Limitations}
While \method{} significantly improves reinforcement learning efficiency, it has a few inherent limitations. First, it is designed for Reinforcement Learning with Verifiable Rewards (RLVR) using objective, binary metrics, making it difficult to apply to subjective human preference alignment (RLHF). Second, the framework relies on static difficulty estimates from a single offline pass. Because the model improves during training, these initial estimates can become outdated, 
requiring re-profiling to maintain optimal compute allocation. 
In our experiments, we profile once at initialization and find 
that the single-pass estimates remain predictive enough to yield 
large compute savings; however, for longer training runs or 
rapidly shifting difficulty distributions, periodic re-profiling 
at epoch boundaries would be a natural extension.

\paragraph*{Ethical Considerations}
By reducing training compute and minimizing wasted rollouts, \method{} lowers the energy consumption and carbon footprint of aligning large language models. This efficiency also democratizes AI development, allowing researchers with fewer hardware resources to train highly capable reasoning models. However, this accessibility raises dual-use risks, potentially making it easier for malicious actors to optimize models for harmful applications. Furthermore, the method's aggressive data filtering, which discards trivial prompts and subsamples unsolvable ones, risks introducing representational harms. If the initial policy's success correlates with specific cultural or linguistic contexts, this filtering could inadvertently amplify existing biases and degrade performance on underrepresented data distributions.

\bibliography{biblio}

\clearpage
\appendix
\onecolumn

\section{Algorithm}
\label{appx:algorithm}

\begin{algorithm}[ht!]
\caption{sorted Group Policy Optimization (\method{})}
\label{alg:sorted_g}
\renewcommand{\algorithmiccomment}[1]{\hfill\textcolor{gray}{// #1}}
\begin{algorithmic}[1]
\Require Training dataset \(\mathcal{D}\), Initial policy \(p_\theta\), Profiling budget \(N=8\), Group sizes \(\mathcal{G} = \{2, 4, 8\}\), Threshold \(t=0.75\)
\vspace{1.5mm}

\Statex \emph{Profiling via Inference-Time Scaling}
\For{each query \(q \in \mathcal{D}\)}
    \State $\{\tau_i\}^N_{i=1} \sim p_\theta(\tau \mid q)$ \Comment{Generate N parallel samples}
    \State \(\hat{p}(q) = \dfrac{n_{\text{profiling}}(q)}{N}\) \Comment{Estimate empirical success rate}
\EndFor
\vspace{1.5mm}

\Statex \emph{Dataset Partitioning and Clustering}
\State \(\mathcal{D}_{\text{trivial}} \gets \{q \in \mathcal{D} \mid \hat{p}(q) > t\}\) \Comment{Remove trivial}
\State \(\mathcal{D}_{\text{unsolved}} \gets \{q \in \mathcal{D} \mid \hat{p}(q) = 0\}\) \Comment{Remove unsolved}
\State \(\mathcal{D}_{\text{learnable}} \gets \{q \in \mathcal{D} \mid 0 < \hat{p}(q) \leq t\}\) \Comment{Use Learnable}
\State Apply bucket map \(b(q)\) to partition \(\mathcal{D}_{\text{learnable}}\) into clusters \(C_2, C_4, C_8\)
\vspace{1.5mm}

\Statex \emph{Curriculum Mixing and Training}
\State \(\tilde{\mathcal{U}} \gets \text{Subsample}(\mathcal{D}_{\text{unsolved}})\) 
\For{each group size \(G \in \{2, 4, 8\}\) in ascending order}
    \State \(\bar{C}_G \gets C_G \cup \tilde{\mathcal{U}}\) \Comment{Mix unsolved queries into the cluster}
    \State Train policy \(p_\theta\) for one epoch on \(\bar{C}_G\) using fixed group size \(G\)
\EndFor
\vspace{1.5mm}

\State \Return Optimized policy \(p_\theta\)
\end{algorithmic}
\end{algorithm}

\clearpage
\section{Training Dynamics}
\label{appx:dynamics}

\begin{figure}[ht!]
\centering
\includegraphics[width=0.85\textwidth]{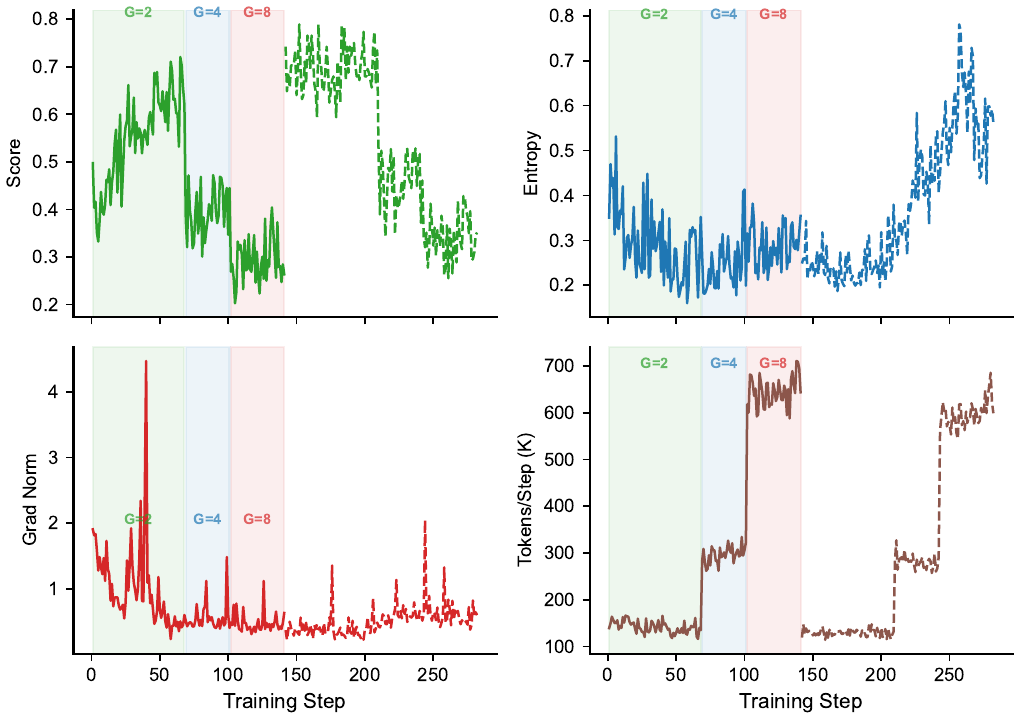}
\caption{Training dynamics across curriculum phases for \method{} on Qwen2.5-Math-7B. Top-left: mean reward score, top-right: policy entropy, bottom-left: gradient norm, bottom-right: tokens per step. Solid: epoch~1, dashed: epoch~2. Phase bands indicate $G{=}2$ (green), $G{=}4$ (blue), $G{=}8$ (red).}
\label{fig:training_dynamics_panel}
\end{figure}

Figure~\ref{fig:training_dynamics_panel} shows four training metrics across the three curriculum phases for Qwen2.5-Math-7B. Solid lines are epoch~1; dashed lines are epoch~2 (same data, fresh rollouts). 

\paragraph{Score} Mean reward score increases within the $G{=}2$ and $G{=}4$ phases and drops at phase transitions as harder queries are introduced. The $G{=}2$ phase peaks at 0.72 and $G{=}4$ at 0.45. The $G{=}8$ phase peaks at 0.40 but ends at 0.26, close to its starting value of 0.27. This flat trajectory suggests that the hardest learnable queries ($\hat{p} = 1/8$) provide limited per-step improvement at this scale, though the ablation in Table~\ref{tab:ablation} confirms that including this phase still contributes to downstream accuracy. In epoch~2, the $G{=}2$ phase starts at 0.74 (vs 0.50 in epoch~1), reflecting retained capability.

\paragraph{Entropy} Policy entropy remains between 0.20 and 0.35 during the $G{=}2$ phase. During $G{=}4$, entropy decreases from 0.27 to 0.20, consistent with the model narrowing its output distribution as it learns to solve medium-difficulty queries more reliably. During the $G{=}8$ phase, entropy rises (0.32$\to$0.36 in epoch~1, 0.45$\to$0.56 in epoch~2), coinciding with the model encountering queries where no profiled success exists.

\paragraph{Gradient norm} Gradient norms peak in the $G{=}2$ phase (avg 1.02) and decrease in later phases (0.56 for $G{=}4$, 0.48 for $G{=}8$). The largest single-step norm (4.47) occurs at training step~1. Later phases produce smaller gradients because the policy has already been updated on easier queries.

\paragraph{Tokens per step} Token count scales approximately with $G$: 142K for $G{=}2$, 297K for $G{=}4$, 645K for $G{=}8$. The slight super-linear scaling (2.1$\times$ and 2.2$\times$ per doubling of $G$) reflects longer average responses on harder queries. The curriculum front-loads the cheapest phase, spending fewer tokens per step when score improvements are largest.

\clearpage
\section{Learning Effectiveness}
\label{appx:learning}

We re-profile the full dataset after one epoch of \method{} training on Qwen2.5-Math-7B to measure how the difficulty distribution changes.

\begin{figure}[ht!]
\centering
\begin{minipage}[b]{0.48\textwidth}
    \centering
    \includegraphics[width=\textwidth]{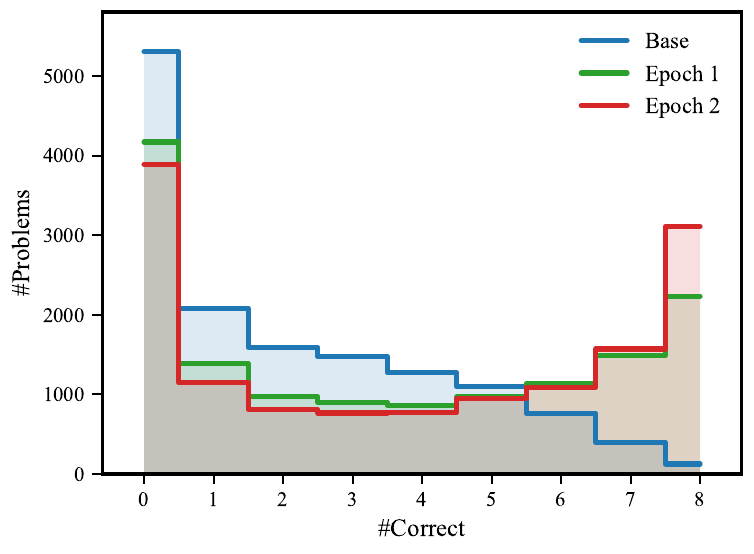}
    \subcaption{Full per-bin distribution before (left) and after (right) training.}
    \label{fig:reprofile_full}
\end{minipage}
\hfill
\begin{minipage}[b]{0.48\textwidth}
    \centering
    \includegraphics[width=\textwidth]{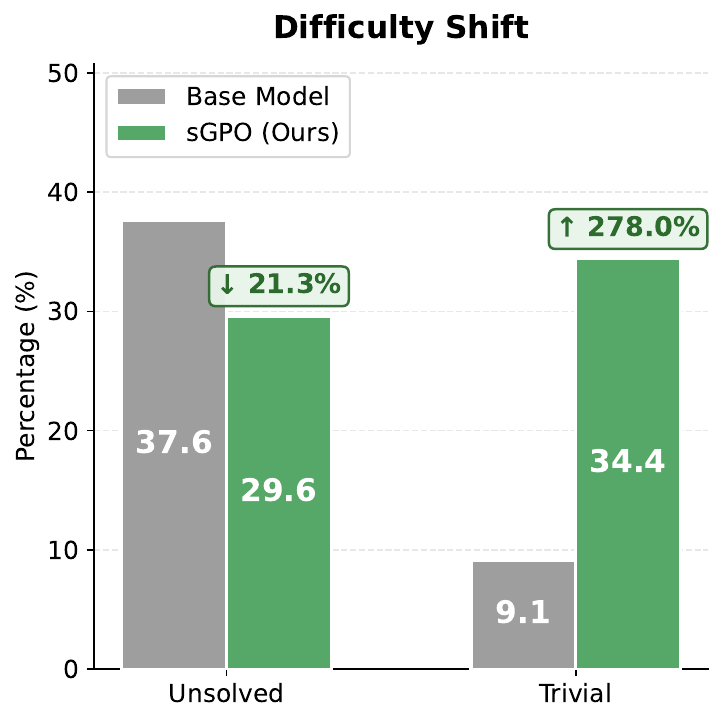}
    \subcaption{Unsolved and trivial fractions before and after training.}
    \label{fig:reprofile_summary}
\end{minipage}
\caption{Difficulty distribution shift after one epoch of \method{} on Qwen2.5-Math-7B. (a) The full profiling distribution across all $\hat{p}$ bins. (b) The two extreme categories: unsolved drops from 37.6\% to 29.6\% ($-$8.0pp), trivial grows from 9.1\% to 34.4\% ($+$25.3pp).}
\label{fig:difficulty_distribution}
\end{figure}

Figure~\ref{fig:reprofile_full} shows the full per-bin distribution before and after training. The base model concentrates at two uninformative extremes: 37.6\% unsolved ($\hat{p}{=}0$) and 9.1\% trivial ($\hat{p} \geq 0.75$). After one epoch, probability mass shifts from the unsolved region toward intermediate and high-correct bins. Figure~\ref{fig:difficulty_distribution}b summarizes the endpoints: the trivial fraction grows by 25.3pp (9.1\% $\to$ 34.4\%) and the unsolved fraction shrinks by 8.0pp (37.6\% $\to$ 29.6\%).

Two findings:

\begin{enumerate}
    \item \textbf{Learnable queries are mastered.} The 25.3pp increase in trivial queries shows that the $G{=}2$ phase converts high-$\hat{p}$ problems from learnable to solved. These queries would produce zero gradient under the same $G^*$ assignment in a second epoch.

    \item \textbf{Unsolved queries become solvable.} 8.0pp of previously unsolved queries ($\hat{p}{=}0$) become solvable after training. The 10\% unsolved mixing ($\alpha{=}10\%$) in each phase provides direct exposure, while capability built on easier queries during the $G{=}2$ and $G{=}4$ phases may also transfer. The ablation in Table~\ref{tab:ablation_mixing} confirms that removing unsolved mixing ($\alpha{=}0\%$) drops accuracy by 2.7pp.
\end{enumerate}

The post-training distribution differs substantially from the pre-training profile, confirming that $\hat G$ assignments become stale over training. This motivates the re-profiling direction discussed in Section~\ref{sec:conclusion}.

\clearpage
\section{Multi-Epoch Training}
\label{appx:multi_epoch}

All main results in this paper use two epochs of \method{} training. This section describes the epoch~2 strategy selection that led to this default.

We evaluated three strategies for continuing into epoch~2 on Qwen2.5-Math-7B, all starting from the same epoch~1 checkpoint:

\begin{enumerate}
    \item \textbf{Stale replay:} Reuse the original profiling assignments ($\hat{p}$ and $\hat G$ from the base model). The curriculum repeats with identical data ordering and group sizes.
    \item \textbf{Re-profiled:} Re-profile the dataset using the epoch~1 checkpoint to obtain updated $\hat{p}$ values, then recompute $\hat G$, filtering, and curriculum ordering.
    \item \textbf{Re-profiled + fresh optimizer:} Same as (2) but reset the optimizer state before epoch~2.
\end{enumerate}

\begin{figure}[ht!]
\centering
\includegraphics[width=0.6\textwidth]{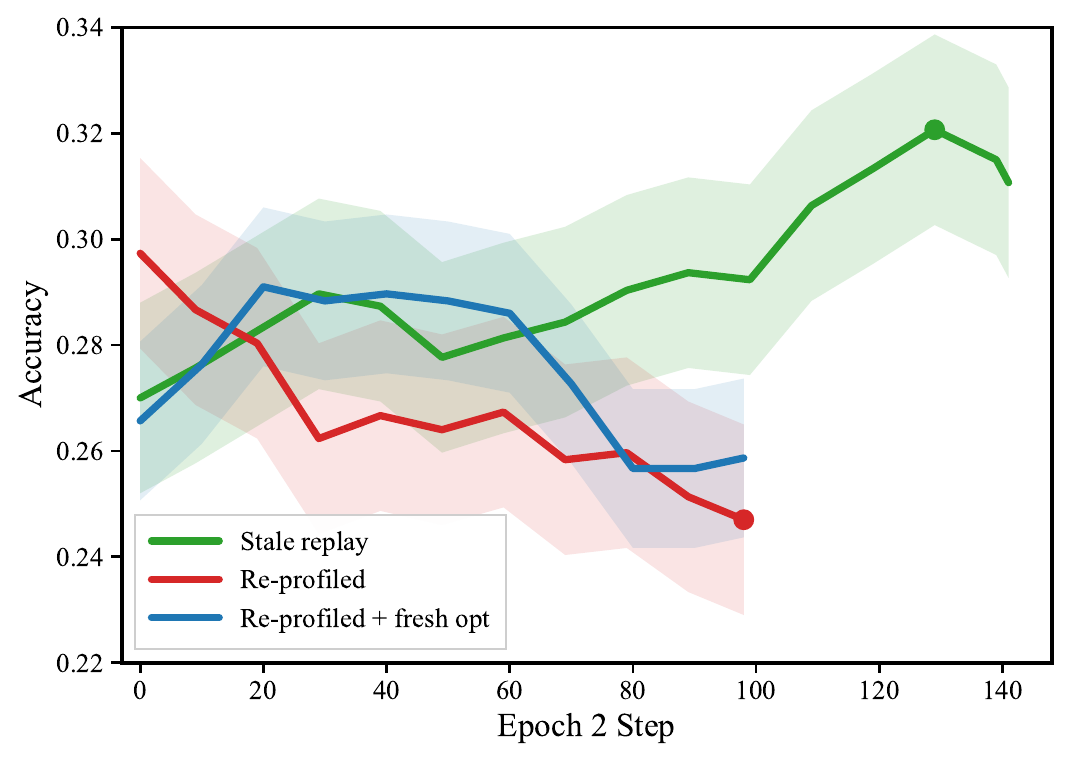}
\caption{Epoch~2 accuracy on Qwen2.5-Math-7B under three continuation strategies. Stale replay (green) improves to ${\sim}32\%$. Both re-profiled variants (red, blue) degrade to ${\sim}25\%$.}
\label{fig:multi_epoch}
\end{figure}

Figure~\ref{fig:multi_epoch} shows the epoch~2 accuracy trajectories. Stale replay improves from 31.0\% to ${\sim}32\%$. Both re-profiled variants degrade to ${\sim}25\%$, below the epoch~1 checkpoint.

Comparing variants (2) and (3) isolates the effect of optimizer state: both use re-profiled assignments, but (3) resets the optimizer. The two perform identically (${\sim}25\%$), indicating that optimizer momentum does not drive the difference. Comparing variants (1) and (2) isolates the effect of assignments: both continue the same optimizer, but (1) uses stale profiling. Stale replay outperforms by ${\sim}7$pp. The profiling assignments, not the optimizer state, determine epoch~2 performance.

\paragraph{Why re-profiling fails}
The $\mathbb{E}[n] \approx 1$ analysis in Section~\ref{subsec:allocation} explains this. After epoch~1, 25.3pp of previously learnable queries reach $\hat{p} \geq 0.75$ under the trained model and are filtered as trivial (Appendix~\ref{appx:learning}). The re-profiled dataset concentrates on hard and unsolved problems. The $G{=}2$ cluster shrinks, eliminating the easy queries that produced the densest gradient signal in epoch~1. Most remaining queries fall into the $\mathbb{E}[n] \approx 0$ dead zone, where groups produce all-incorrect rollouts and zero advantage.

\paragraph{Why stale replay works}
Consider a query that was $\hat{p} = 1/8$ ($G{=}8$, $\mathbb{E}[n]{=}1$) at profiling time and now has true success probability $p \approx 3/8$ after epoch~1. Its stale assignment $G{=}8$ yields $\mathbb{E}[n] = 3$: suboptimal but still producing non-zero advantage, because the group contains a mix of correct and incorrect rollouts. The advantage landscape (Figure~\ref{fig:g_landscape}) has a broad ridge around $\mathbb{E}[n]{=}1$, and $\mathbb{E}[n]{=}3$ remains in the productive region. Re-profiling, by contrast, concentrates the entire distribution near $\mathbb{E}[n]{=}0$.

\paragraph{Implication}
In this setting, retaining difficulty diversity outperforms recalibrating $\hat G$ to the model's current capability. A training distribution with a range of $\mathbb{E}[n]$ values (some at 1, some at 2--3, some near 0) produces gradient on average, while a distribution concentrated at $\mathbb{E}[n] \approx 0$ produces nothing. Based on this finding, all main results use stale replay for epoch~2.

\clearpage
\section{Self-Consistency vs.\ Verified Profiling}
\label{appx:self_consistency}

A natural question is whether \method{}'s profiling pass requires ground-truth verification, or whether a cheaper signal such as self-consistency (SC~\cite{wang2022self}) could substitute. SC measures agreement among sampled responses without checking correctness: if all $N$ responses agree, the query is classified as easy regardless of whether the consensus answer is right.

\begin{figure}[ht!]
\centering
\includegraphics[width=0.7\textwidth]{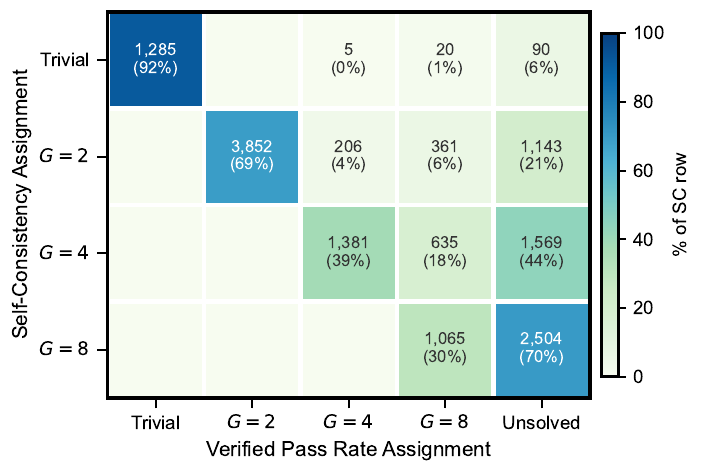}
\caption{Confusion matrix between self-consistency (SC) and verified pass-rate difficulty assignments on DAPO-Math-17k (Qwen2.5-Math-7B, $N{=}8$). Each cell shows the count and row-normalized percentage. SC never classifies queries as unsolved (no ``Unsolved'' row exists) because it cannot distinguish consistent correctness from consistent error. The ``Unsolved'' column intensifies downward: 21\% of SC's $G{=}2$ queries, 44\% of $G{=}4$, and 70\% of $G{=}8$ are actually unsolved.}
\label{fig:sc_confusion}
\end{figure}

Figure~\ref{fig:sc_confusion} cross-tabulates SC-based and pass-rate-based difficulty assignments. SC agrees with verified profiling for trivial queries (92\% correct) and $G{=}2$ queries (69\% correct). The disagreement concentrates in the ``Unsolved'' column: 1,143 queries that SC assigns to $G{=}2$, 1,569 to $G{=}4$, and 2,504 to $G{=}8$ are in fact unsolved ($\hat{p}{=}0$). In these cases, the model produces consistent but incorrect answers, and SC interprets the agreement as evidence of capability.

SC-based profiling would allocate training compute to these queries, generating zero gradient signal because no correct response exists in the rollout group. Verified profiling avoids this: queries with $\hat{p}{=}0$ are identified and filtered (or subsampled at $\alpha{=}10\%$). The cost of verification is low when rewards are binary and automatically checkable, the standard setting in RLVR.

\clearpage
\section{Uniform Group Scaling}
\label{appx:g16}

\begin{figure}[ht!]
\centering
\includegraphics[width=0.7\textwidth]{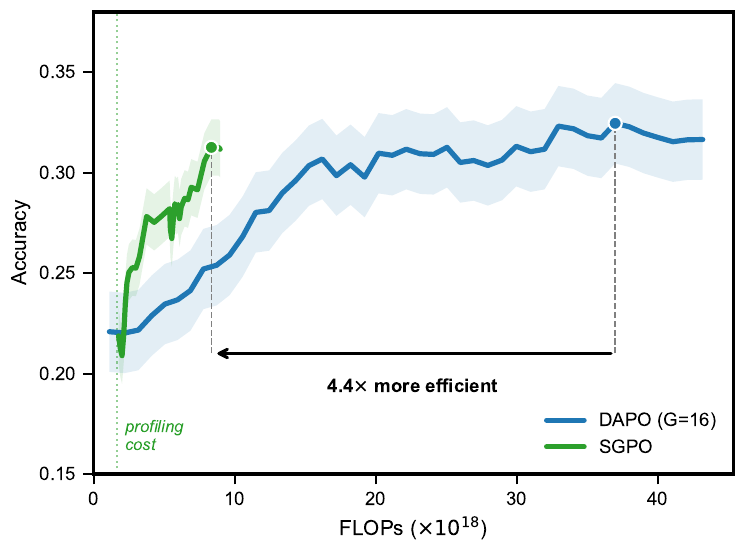}
\caption{Accuracy vs.\ cumulative FLOPs for \method{} and DAPO ($G{=}16$) on Qwen2.5-Math-7B. \method{} reaches comparable peak accuracy 4.4$\times$ faster.}
\label{fig:flop_g16}
\end{figure}

Does increasing the uniform group size close the gap with adaptive allocation? Figure~\ref{fig:flop_g16} compares \method{} ($G \in \{2,4,8\}$, max group size 8) against DAPO with $G{=}16$ on Qwen2.5-Math-7B.

DAPO $G{=}16$ peaks at ${\sim}32\%$, 1pp above \method{}'s $31.0\%$, but consumes ${\sim}36$~EF, which is 4.4$\times$ more compute than \method{}'s 8.9~EF. Doubling the group size from $G{=}8$ to $G{=}16$ increases DAPO's total FLOPs by 61\% (22.3 $\to$ 36~EF) while improving peak accuracy by only ${\sim}2$pp. Most of the additional rollouts land on queries where the model already succeeds or consistently fails, producing zero advantage.

\method{} matches this accuracy with a maximum group size of 8 by allocating $G{=}8$ only to the hardest learnable queries ($\hat{p} = 1/8$). Easier queries receive $G{=}2$ or $G{=}4$, avoiding the redundant rollouts that dominate uniform $G{=}16$ training.

\clearpage
\section{Experimental Details}
\label{appx:details}

\subsection{Datasets and Models}

Tables~\ref{tab:train_data}, \ref{tab:eval_data}, and \ref{tab:models}
summarize the datasets, benchmarks, and models used in this work.

\begin{table}[ht!]
\centering
\begin{tabular}{lcc}
\toprule
Dataset & Problems & HuggingFace Url \\
\midrule
DAPO-Math-17k & 14,116 & \texttt{BytedTsinghua-SIA/DAPO-Math-17k} \\
SciKnowEval & 3901 & \texttt{hicai-zju/SciKnowEval} \\
\bottomrule
\end{tabular}
\caption{Training datasets. DAPO-Math-17k contains 14,116 English problems after deduplication. SciKnowEval uses a 90/10 train-test split of the L3 (reasoning) subset.}
\label{tab:train_data}
\end{table}

\begin{table}[ht!]
\centering
\begin{tabular}{lcc}
\toprule
Benchmark & Problems & HuggingFace Url  \\
\midrule
AIME 2024 & 30 & \texttt{Maxwell-Jia/AIME\_2024} \\
AIME 2025 & 30 & \texttt{MathArena/aime\_2025} \\
AIME 2026 & 30 & \texttt{MathArena/aime\_2026} \\
HMMT Feb 2025 & 30 & \texttt{MathArena/hmmt\_feb\_2025} \\
HMMT Feb 2026 & 33 & \texttt{MathArena/hmmt\_feb\_2026} \\
SciKnowEval-L3 Chemistry & 210 & \texttt{hicai-zju/SciKnowEval} \\
SciKnowEval-L3 Physics & 80 & \texttt{hicai-zju/SciKnowEval} \\
SciKnowEval-L3 Biology & 50 & \texttt{hicai-zju/SciKnowEval} \\
SciKnowEval-L3 Material Sciences & 94 & \texttt{hicai-zju/SciKnowEval} \\
\bottomrule
\end{tabular}
\caption{Evaluation benchmarks. SciKnowEval counts reflect the 10\% test split.}
\label{tab:eval_data}
\end{table}

\begin{table}[ht!]
\centering
\begin{tabular}{lcc}
\toprule
Model & Params & HuggingFace Path  \\
\midrule
Qwen2.5-Math-1.5B~\citep{yang2024qwen2} & 1.5B & \texttt{Qwen/Qwen2.5-Math-1.5B} \\
Qwen2.5-Math-7B~\citep{yang2024qwen2} & 7B & \texttt{Qwen/Qwen2.5-Math-7B} \\
Qwen3-4B-Instruct-2507~\citep{yang2025qwen3} & 4B & \texttt{Qwen/Qwen3-4B-Instruct-2507} \\
\bottomrule
\end{tabular}
\caption{Models used in our experiments. Qwen2.5-Math models are base (non-instruct) checkpoints. Qwen3-4B is instruction-tuned.}
\label{tab:models}
\end{table}

\subsection{Prompt}

\begin{figure}[ht!]
\begin{tcolorbox}[colback=white, colframe=black, fonttitle=\bfseries, title=Prompt Template]
\small
\texttt{<|im\_start|>system}\\
Please reason step by step, and put your final answer within \textbackslash boxed\{\}.\\
\texttt{<|im\_end|>}\\[4pt]
\texttt{<|im\_start|>user}\\
Please solve the following math problem: \textit{\{\{Question\}\}}. The assistant first thinks about the reasoning process step by step and then provides the user with the answer. Return the final answer in \textbackslash boxed\{\} tags, for example \textbackslash boxed\{1\}. Let's solve this step by step.\\
\texttt{<|im\_end|>}\\[4pt]
\texttt{<|im\_start|>assistant}
\end{tcolorbox}
\caption{Prompt template used for training and evaluation. The system message is auto-injected by the Qwen2.5-Math tokenizer. The user message wraps the problem in the instruction format.}
\label{fig:prompt_template}
\end{figure}

\clearpage
\subsection{Training and Evaluation Setup}
\label{appx:training-detail}
All experiments use DAPO as the underlying RL optimizer, trained on $8 \times$ H100 GPUs. \method{} adds profiling, filtering, rollout allocation, and curriculum ordering on top of the standard DAPO pipeline; all other hyperparameters (learning rate, KL penalty, clipping) remain unchanged.

The profiling budget is fixed at $N=8$ samples per query. We define trivial queries as
those with $\hat{p}(q) > 0.75$, unsolved queries as those with $\hat{p}(q)=0$, and
learnable queries as those with $0 < \hat{p}(q) \leq 0.75$. Learnable queries are
bucketed into rollout groups $G \in \{2,4,8\}$, and we mix an $\alpha=10\%$ subsample
of unsolved queries into each curriculum phase, training sequentially over
$\bar{C}_2 \rightarrow \bar{C}_4 \rightarrow \bar{C}_8$.

Evaluation uses avg@16 accuracy: for each query, we generate 16 independent samples at temperature $1.0$ with top-$p=0.7$ and report the fraction of samples that are correct, averaged across all queries.

\subsection{FLOP Accounting}
\label{app:flop_accounting}

We report compute in ExaFLOPs ($\times 10^{18}$), where $P$ denotes the number of model parameters. We follow the standard approximation that a transformer forward pass costs ${\sim}2P$ FLOPs per token~\citep{kaplan2020scaling,hoffmann2022training}.

\paragraph{Profiling cost.} The profiling pass performs inference only (autoregressive generation, no gradient computation). Each token costs $2P$ FLOPs:
\[
\mathrm{FLOPs}_{\mathrm{profiling}} = 2P \cdot T_{\mathrm{profiling}},
\]
where $T_{\mathrm{profiling}}$ is the total number of tokens generated across all $N{=}8$ samples for all queries.

\paragraph{Training cost.} Each GRPO-style training step processes generated tokens through multiple stages. For DAPO~\citep{yu2025dapo}, which uses a frozen reference model for KL regularization, the per-token cost breaks down as:

\begin{enumerate}
    \item \textbf{Rollout generation} (actor, forward only): $2P$
    \item \textbf{Reference log-probabilities} (reference model, forward only): $2P$
    \item \textbf{Actor update} (forward + backward): $2P + 4P = 6P$
\end{enumerate}

\noindent This totals ${\sim}10P$ per token. In practice, implementation overhead (gradient accumulation, KL computation, advantage normalization) and framework-level inefficiencies raise the effective cost. We use $12P$ as a conservative upper bound, consistent with estimates in prior work~\citep{li2025knapsack}. Methods that eliminate the reference model (e.g., Dr.~GRPO~\citep{liu2025understanding}) reduce this to ${\sim}8P$, which would further increase \method{}'s relative advantage.

\[
\mathrm{FLOPs}_{\mathrm{train}} = 12P \cdot T_{\mathrm{train}}
\]

\paragraph{Total cost.}
\[
\mathrm{FLOPs}_{\mathrm{total}} =
\mathrm{FLOPs}_{\mathrm{profiling}} + \mathrm{FLOPs}_{\mathrm{train}}
= 2P \cdot T_{\mathrm{profiling}} + 12P \cdot T_{\mathrm{train}}
\]

The $6\times$ gap between inference ($2P$) and training ($12P$) per token is the cost asymmetry that \method{} exploits: profiling tokens are cheap, training tokens are expensive.

\end{document}